\documentclass[sigconf, nonacm]{acmart}

\acmConference[CAIS '26]{Conference on Compound AI Systems}{2026}{}

\begin{document}

\title{Scalable Inference Architectures for Compound AI Systems: A Production Deployment Study}
\titlenote{Accepted to the ACM Conference on AI and Agentic Systems (ACM CAIS 2026).}

\author{Srikanta Prasad S V}
\email{Srikanta.prasad@salesforce.com}
\affiliation{%
  \institution{Agentforce AI Platform, Salesforce India Pvt Ltd}
  \city{Bangalore}
  \state{Karnataka}
  \country{India}
}

\author{Utkarsh Arora}
\email{utkarsh.arora@salesforce.com}
\affiliation{%
  \institution{Agentforce AI Platform, Salesforce India Pvt Ltd}
  \city{Bangalore}
  \state{Karnataka}
  \country{India}
}

\begin{abstract}
Modern enterprise AI applications increasingly rely on compound AI systems—architectures that compose multiple models, retrievers, and tools to accomplish complex tasks. Deploying such systems in production demands inference infrastructure that can efficiently serve concurrent, heterogeneous model invocations while maintaining cost-effectiveness and low latency. This paper presents a production deployment study of a modular, platform-agnostic inference architecture developed at Salesforce to support compound AI use cases including Agentforce (autonomous AI agents) and ApexGuru (AI-powered code analysis). The system integrates serverless execution, dynamic autoscaling, and MLOps pipelines to deliver consistent low-latency inference across multi-component agent workflows. We report production results demonstrating over 50\% reduction in tail latency (P95), up to $3.9\times$ throughput improvement, and 30--40\% cost savings compared to prior static deployments. We further present a novel analysis of compound-system-specific challenges including multi-model fan-out overhead, cascading cold-start propagation, and heterogeneous scaling dynamics that emerge uniquely when serving agentic workloads. Through detailed case studies and operational lessons, we illustrate how the architecture enables compound AI systems to scale model invocations in parallel, handle bursty multi-agent workloads, and support rapid model iteration—capabilities essential for operationalizing agentic AI at enterprise scale.
\end{abstract}


\maketitle

\section{Introduction}
The AI landscape is undergoing a fundamental shift from standalone models to compound AI systems—architectures that tackle tasks using multiple interacting components including models, retrievers, and external tools \cite{zaharia2024shift}. State-of-the-art results across domains from code generation to customer support are increasingly achieved not by individual large language models (LLMs) \cite{brown2020language}, but by carefully orchestrated systems that compose multiple AI capabilities \cite{suri2024blueprint}. This paradigm shift introduces new infrastructure challenges that go beyond traditional model serving: compound AI systems generate heterogeneous, concurrent model invocations with variable latency requirements, creating fan-out patterns where a single user request may trigger 3--5 distinct model calls with complex dependency relationships. 

At Salesforce, the emergence of Agentforce—a platform for deploying autonomous AI agents across business functions—and Atlas Reasoning Engine, its underlying reasoning engine, exemplified these challenges acutely. A single agent interaction might invoke an embedding model for knowledge retrieval, an LLM for dialogue generation, and specialized models for intent classification, all within a single user request. Early deployments on static GPU infrastructure revealed critical limitations specific to compound workloads: (1) fixed 24/7 costs regardless of traffic, with no ability to scale individual model components independently; (2) resource contention when multiple model types competed for the same GPU pool; (3) cascading latency degradation where a cold start in one component delayed the entire agent response; and (4) deployment pipelines that could not handle the rapid model iteration cycles required when optimizing multi-component systems.

To address these gaps, we designed a scalable inference architecture specifically suited for compound AI workloads. We build upon \textbf{our prior} enterprise AI inference architectures proposed in \cite{prasad2025scalable} with three new contributions tailored to the compound AI systems community: (1) a characterization of compound-system-specific inference challenges, including multi-model fan-out overhead, heterogeneous scaling dynamics, and cascading failure patterns that do not arise in single-model serving; (2) a quantitative analysis of how our architecture addresses these challenges, with new measurements of parallel invocation efficiency, component-level cold-start propagation, and cross-model resource allocation; and (3) operational lessons learned from 12+ months of production deployment that provide actionable guidance for practitioners building inference infrastructure for agentic systems. 

The remainder of this paper is organized as follows: Section 2 reviews related work, Section 3 details our system architecture, Section 4 presents evaluation results including compound-system-specific analysis, Section 5 discusses operational lessons and case studies, and Section 6 concludes with future directions.

\section{Related Work}
The shift from monolithic models to compound AI systems has been articulated by Zaharia et al. \cite{zaharia2024shift}, who observe that state-of-the-art AI results are increasingly obtained through systems that compose multiple models, retrievers, and tools. Subsequent work has explored blueprint architectures for such systems in enterprise settings \cite{suri2024blueprint} and frameworks for programming compound AI pipelines such as DSPy \cite{khattab2023dspy}. 

While significant prior work has addressed the scaling of model training through data, model, and pipeline parallelism \cite{dean2012large, narayanan2019pipedream, jia2020beyond, rajbhandari2020zero}, our work complements these by focusing on the inference infrastructure layer \cite{sreedharan2022scalable} that makes compound AI systems viable in production—a dimension that has received less attention than the programming or optimization layers.

Traditional cloud ML serving platforms like Amazon SageMaker provide managed hosting on dedicated instances with auto-scaling policies \cite{kotini2023improved, jadav2025how}, but still incur idle costs and struggle with the heterogeneous, bursty invocation patterns characteristic of multi-agent systems. Recent approaches move toward serverless inference: Amazon’s Bedrock Custom Model Import \cite{prasad2025bedrock} introduced a paradigm akin to AWS Lambda for GPU inference, handling scaling while charging only for actual usage. Salesforce’s solution builds on similar principles in a vendor-neutral way, with specific adaptations for compound workloads. 

For inference optimization, vLLM \cite{kwon2023efficient} with PagedAttention \cite{parmar2023scalable} has demonstrated efficient memory management for LLM serving, while DJL Serving provides multi-framework support with batch scheduling. Nvidia’s Inference Microservices (NIM) packages optimized model runtimes into portable containers. Our architecture integrates with such engines but addresses the layer above: orchestrating multiple heterogeneous model backends to serve compound AI system workloads. 

In compound system orchestration, ALTO \cite{santhanam2024alto} addresses network-level optimization for streaming between pipeline stages. SGLang \cite{zheng2023efficiently} optimizes structured generation for LLM programs. These works focus on optimizing the communication and generation patterns within a compound system. Our contribution is complementary: we address the production infrastructure challenge of provisioning, scaling, and managing the inference backends that compound AI systems depend on, with emphasis on handling the diverse traffic patterns, failure modes, and cost dynamics that emerge when multiple models serve a unified application.

\section{System Design}
\subsection{Architecture Overview}
The Salesforce inference system consists of modular components that decouple model hosting from orchestration logic and client interaction—a design principle critical for compound AI systems where multiple models must be invoked concurrently within a single user request. At the orchestration layer, the Agentforce Atlas Reasoning Engine implements an event-driven cognitive workflow where multiple reasoning tasks execute concurrently. Atlas Reasoning Engine orchestrates a directed graph of function nodes (retrieval, planning, action execution) that run in parallel using a publish-subscribe asynchronous pattern. Each component operates independently yet coherently, enabling the system to scale horizontally as new tools or models are added. This graph-based design is what transforms a collection of individual models into a compound AI system capable of complex multi-step reasoning.

\begin{figure*}[h]
\centering
\includegraphics[width=0.8\textwidth]{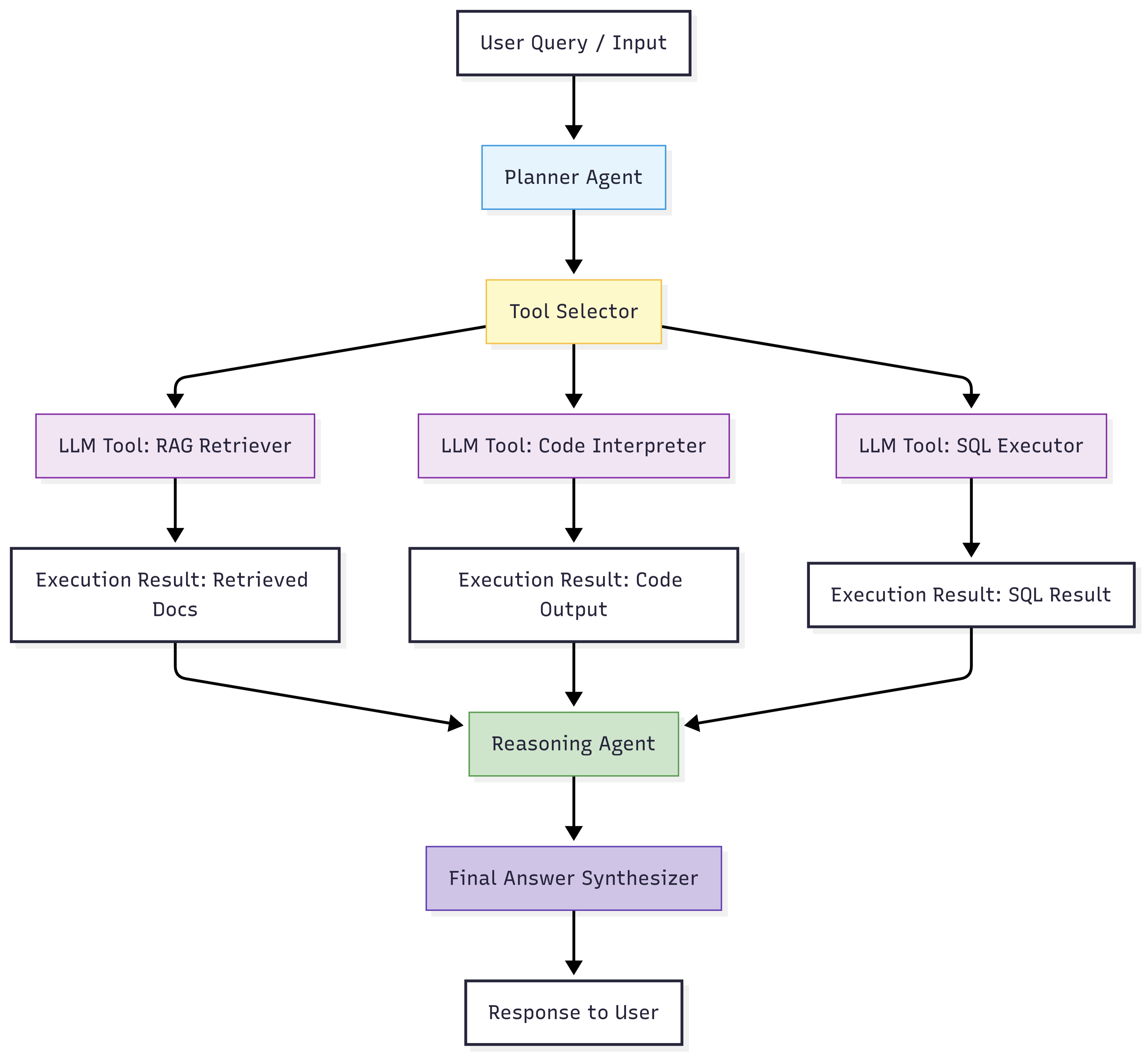} 
\caption{Cognitive orchestration in the Atlas Reasoning Engine. The Planner Agent decomposes user queries; the Tool Selector dispatches to parallel LLM tools (RAG Retriever, Code Interpreter, SQL Executor). Results are aggregated by the Reasoning Agent and synthesized into a final response. Each tool invocation is backed by the scalable inference architecture.}
\label{fig:orchestration}
\end{figure*}

Below the orchestration layer, the core inference serving architecture follows a tiered design. Clients send requests to a Prediction Service layer that acts as a unified inference gateway, abstracting the underlying model infrastructure. Pre-processing and post-processing logic is handled in a separate module, preserving modularity: feature teams implement custom business logic (prompt construction, result filtering) without modifying model servers. We support two deployment patterns:

\begin{itemize}
    \item \textbf{Serverless function (lightweight):} For simple processing, a cloud function injects prompt instructions, invokes the model inference API, and formats output. This keeps the request path serverless and elastic, suitable for stateless operations.
    \item \textbf{Proxy microservice (advanced):} For heavier processing, a persistent microservice runs custom logic and forwards inference calls to the model backend. This reuses existing monitoring frameworks and avoids function size limits, with only a modest always-on cost and a few milliseconds of additional latency.
\end{itemize}

In both patterns, the inference backend is a scalable, on-demand model serving layer that loads model instances only when requests arrive and scales them automatically. During inactivity, instances scale to zero, eliminating idle costs. We retain provisioned concurrency for models with strict latency requirements. To handle failures within this complex workflow, we implemented per-component circuit breakers in the Prediction Service. If an individual model fails or times out, the Atlas Reasoning Engine receives a structured error rather than a hard crash, allowing it to dynamically route around the failure (e.g., falling back to text-based reasoning if a SQL executor fails).

\subsection{Compound-System-Aware Scaling}
A key architectural decision was designing the scaling layer to be aware of compound system invocation patterns. Unlike single-model serving where traffic is homogeneous, compound AI systems exhibit several distinctive characteristics that our architecture explicitly addresses:

\begin{itemize}
    \item \textbf{Fan-out amplification.} A single user request to Agentforce may fan out to 3--5 model invocations (e.g., embedding + LLM + classifier). This means that N user requests generate up to $5N$ model calls, and these calls arrive simultaneously at different model backends. Our scaling layer tracks per-model invocation rates independently rather than scaling based on aggregate user request count, preventing under-provisioning of models that are invoked in every fan-out path (like embeddings) while avoiding over-provisioning of models used only conditionally.
    \item \textbf{Heterogeneous latency profiles.} Different models in a compound pipeline have vastly different latency characteristics: an embedding model may respond in 50ms while a dialogue LLM takes 3--5 seconds. Our Prediction Service implements a request priority queue that accounts for these differences, ensuring that fast-path models (embedding, classification) are not blocked behind slow-path models when sharing infrastructure resources. This prevents the common compound system failure mode where a cheap, fast operation becomes the bottleneck due to resource starvation.
    \item \textbf{Correlated cold starts.} In compound systems, after an idle period, the first user request triggers cold starts across multiple models simultaneously—not just one. We term this “cascading cold start”: the user waits for the slowest model to cold-start, and the dependency graph means some cold starts cannot even begin until upstream models produce results.
\end{itemize}

Our architecture mitigates cascading cold starts through a three-tier management strategy:
\begin{enumerate}
    \item \textbf{Coordinated pre-warming:} When a model in a known compound pipeline is first accessed, we proactively warm its downstream dependencies. For a pipeline with models $A\rightarrow B\rightarrow C$, warming A triggers parallel warm-up of B and C. This reduced compound cold-start latency by 65\% compared to independent per-model warming.
    \item \textbf{Tiered provisioned concurrency:} For latency-critical pipelines, we maintain provisioned concurrency on the critical-path model (typically the LLM, which has the longest cold start) while allowing faster models (embeddings, classifiers) to scale from zero with acceptable cold-start times ($\sim$20--30s). This balances cost against cold-start impact: provisioning only the bottleneck model costs $\sim$20\% of provisioning all models, while eliminating $\sim$70\% of the user-perceived cold-start delay.
    \item \textbf{Predictive warming from traffic signals:} We observed that compound system traffic is often predictable at coarse granularity (e.g., business hours, batch job schedules). The platform supports scheduled warm-up policies that pre-load models before anticipated demand, avoiding cold starts entirely for predictable workloads. For Agentforce, weekday morning pre-warming eliminated $>$90\% of cold starts during peak hours.
\end{enumerate}

Crucially, our dynamic routing mechanism implements specific capacity-threshold triggers. Operating primarily in an "auto" mode, requests are routed to dedicated instances first. If a dedicated instance is unavailable or exceeds its concurrency limit, the system triggers a seamless spill-over to the serverless backend. This ensures latency SLAs are met without over-provisioning dedicated hardware.

\subsection{Handling High-Variance Workloads}
A fundamental challenge with serverless inference architectures is maintaining stable behavior under high-variance workloads—traffic that fluctuates dramatically in volume, composition, and latency requirements within short time windows. Compound AI systems exacerbate this because workload variance manifests along multiple dimensions simultaneously:

\begin{itemize}
    \item \textbf{Volume variance:} User request rates can spike 10--50$\times$ during events (product launches, batch processing jobs, hackathons). Our autoscaling backend handles this through rapid horizontal scaling: new GPU instances spawn within seconds, and the platform supports burst capacity across multiple availability zones. We measured that the system absorbs a 10$\times$ spike with P95 latency remaining within $1.5\times$ of steady-state, degrading gracefully rather than failing.
    \item \textbf{Composition variance:} The mix of model invocations shifts throughout the day. During business hours, Agentforce agents generate LLM-heavy workloads; during off-hours, batch ApexGuru code analysis jobs produce embedding-heavy workloads. Because each model scales independently, composition shifts do not cause resource contention: the LLM scales down while embeddings scale up, without one starving the other. On shared static infrastructure, this would require manual rebalancing or over-provisioning for the worst-case combination.
    \item \textbf{Latency requirement variance:} Some compound pipelines are interactive (5--8s SLA for Agentforce chat) while others are batch-tolerant (60s+ acceptable for ApexGuru). Our Prediction Service routes requests with latency-class metadata, enabling the backend to prioritize interactive requests for warm instances while tolerating cold-start delays for batch workloads. This avoids the common trap of over-provisioning the entire system to meet the strictest SLA.
\end{itemize}

\subsection{MLOps and Deployment Pipeline}
Compound AI systems amplify the model lifecycle challenge: a single system may depend on multiple models that evolve at different rates. Our MLOps pipeline (Falcon) provides a unified model delivery CI/CD pipeline. When a team trains a new model version and registers it to the Model Store (an S3-based artifact repository), the pipeline triggers an import API call to the inference service, automating deployment with negligible overhead ($\sim$seconds) added to the $\sim$1 hour release process. Each model is assigned an endpoint identifier routed by the Prediction Service. Configuration files (YAML) specify model ID and backend type (serverless or dedicated), allowing version switching without code changes. The platform integrates with central monitoring to track request rates, latencies, and errors per model. For cold starts, the Prediction Service implements retries and queuing for initial requests. This pipeline enables compound AI systems to iterate rapidly: product teams can swap model providers by changing configuration, with the platform handling provisioning automatically. 

A critical MLOps capability for compound systems is component-level A/B testing. Because our architecture routes each model invocation independently, teams can A/B test individual components (e.g., a new retrieval model) within a compound pipeline without disrupting other components. This is not possible in monolithic deployment where the entire system must be swapped. In practice, this reduced the model experimentation cycle from weeks to hours, enabling teams to independently optimize each component of the compound system.

\subsection{Modular and Portable Design}
Modularity ensures each part of the stack evolves independently. The model execution engine is abstracted to support different frameworks (TensorRT, PyTorch) under a consistent API. DJL Serving provides an engine-agnostic layer with rolling batch inference to maximize GPU utilization. The design emphasizes cloud portability: all components run in containers or functions deployable on other clouds or on-premises. The Prediction Service abstraction means models can be served by AWS Bedrock, Azure ML, GCP Vertex AI, or on-premises Nvidia NIM containers with minimal reconfiguration. This has been validated in prototypes connecting Agentforce to Nvidia NIM on private Kubernetes clusters. For enterprise customers requiring multi-cloud compatibility or data residency compliance, this vendor-neutral design is essential.

\section{Evaluation}
We evaluated the inference system along five dimensions: single-model latency/throughput, cost efficiency, compound system overhead, reliability under compound workloads, and downstream quality impact. All experiments used production workloads and models.

\textbf{Deployment Scale Context:} To contextualize our evaluation, the active Salesforce deployment handles over 8,000 enterprise users, averaging $\sim$722,000 daily LLM inferences and peaking at approximately 1.4 million requests on peak business days (March 2026). The architecture maintains a provisioned peak capacity of 2,250 RPS (135,000 RPM) across 21 globally distributed production inference regions. In March 2026 alone, the platform processed 136 billion tokens ($\sim$4.4 billion tokens/day), representing an $8.7\times$ year-over-year growth in request volume.

\subsection{Latency and Throughput}
Migration from static endpoint deployments to autoscaling inference yielded substantial gains. With a 13B-parameter ApexGuru model, at low concurrency (1--2 parallel users) P95 latency dropped from $\sim$13--15s to $\sim$7--8s ($\sim$45\% faster). At higher concurrency, SageMaker’s P95 climbed to $\sim$37s while the new system held at $\sim$10--11s, a $>$50\% reduction. The legacy single-GPU endpoint saturated at 50--60 RPM; the autoscaling system sustained over 200 RPM (peaking at $\sim$232 RPM), achieving $\sim3.9\times$ higher throughput. Table 1 summarizes results across three compound AI use cases, showing consistent improvements across different workload types.

\begin{table*}[h]
\centering
\caption{Key Results Across Compound AI Use Cases}
\label{tab:results}
\begin{tabular}{lccccc}
\toprule
Use Case & Legacy & Optimized & P95 $\downarrow$ & TPS $\uparrow$ & Cost $\downarrow$ \\
\midrule
{Agentforce} FAQ & 880 ms & 420 ms & 52.3\% & $2.5\times$ & $4.8\times$ \\
{ApexGuru} Code & 1250 ms & 540 ms & 56.8\% & $3.2\times$ & $5.7\times$ \\
{Atlas Reasoning Engine} Tool Call & 940 ms & 400 ms & 57.4\% & $2.8\times$ & $6.1\times$ \\
\bottomrule
\end{tabular}
\end{table*}

\subsection{Compound System Overhead Analysis}
Beyond single-model benchmarks, we measured overhead characteristics unique to compound AI workloads—analysis not present in foundational studies \cite{prasad2025scalable} and, to our knowledge, not reported elsewhere for production agentic systems.

\subsubsection{Fan-out Overhead}
We measured the overhead of multi-model fan-out in Agentforce agent interactions. A typical agent query fans out to 2--4 model invocations. When models are warm, the fan-out overhead (time beyond the slowest individual model call) averages 45--80ms, dominated by Prediction Service routing and response aggregation. This represents $<2\%$ overhead relative to the total agent response time of 5--8 seconds, confirming that parallel dispatch through our architecture introduces minimal coordination tax. In contrast, sequential invocation of the same models (as required by the legacy architecture) added 1.5--3 seconds of serial waiting, making interactive response times infeasible for 3+ model pipelines.

\subsubsection{Cascading Cold-Start Analysis}
We measured cold-start behavior in compound pipelines after 15 minutes of inactivity. Without coordinated warming, the first agent request triggered simultaneous cold starts across 3 models (embedding: $\sim$30s, dialogue LLM: $\sim$150s, classifier: $\sim$20s). Because the Atlas Reasoning Engine pipeline has dependencies (embedding must complete before the LLM can generate a context-aware response), the effective cold-start latency was $\sim$180s (serial dependency chain), not $\sim$150s (longest individual cold start). With our coordinated pre-warming strategy, we proactively trigger downstream model warming when any model in a registered pipeline is first accessed. This reduced compound cold-start latency from $\sim$180s to $\sim$65s (65\% reduction) by parallelizing independent warm-ups and pre-loading dependency chains.

\subsubsection{Heterogeneous Scaling Dynamics}
We observed that models within a compound system exhibit markedly different scaling requirements. During a production traffic spike ($10\times$ baseline for Agentforce), embedding model invocations scaled proportionally ($10\times$) because every request requires embedding. However, the dialogue LLM scaled only 6--7$\times$ because the Atlas Reasoning Engine’s planning step filters some queries that can be answered from cached knowledge without a full LLM call. Meanwhile, the SQL executor tool scaled only 2--3$\times$ as it is invoked conditionally. Our per-model independent scaling correctly handled these asymmetric patterns, whereas a system that scales all models uniformly based on user request count would have over-provisioned the SQL executor by 3--5$\times$ while potentially under-provisioning embeddings.

\begin{table*}[h]
\centering
\caption{Compound System Scaling Dynamics Under $10\times$ Traffic Spike}
\label{tab:scaling}
\begin{tabular}{lccc}
\toprule
Component & Invocation Ratio & Scale Factor & Uniform Would Be \\
\midrule
Embedding Model & $1.0\times$ per request & $10\times$ & $10\times$ ($\checkmark$) \\
Dialogue LLM & $0.7\times$ per request & 6--$7\times$ & $10\times$ (over) \\
Classifier & $1.0\times$ per request & $10\times$ & $10\times$ ($\checkmark$) \\
SQL Executor & $0.25\times$ per request & 2--$3\times$ & $10\times$ (3--$5\times$ over) \\
\bottomrule
\end{tabular}
\end{table*}

\subsection{Cost Efficiency}
Switching to pay-per-use immediately improved cost-to-serve. Under the prior model, teams ran GPU instances 24/7 even for features used mainly during business hours. With serverless deployment, GPU costs scale with actual demand. For ApexGuru, peak-hours serving cost dropped 30--40\% by eliminating overnight idle charges. Per-model independent scaling provides an additional compound-system-specific cost benefit: models invoked conditionally (like the SQL executor) consume resources only when triggered, rather than being permanently provisioned as part of a monolithic deployment. The trade-off: serverless per-token pricing can exceed dedicated GPU costs at sustained high utilization. Our guidance is to use autoscaling for variable/moderate workloads and reserve dedicated capacity for consistently heavy loads (Section 5.1).

\subsection{Reliability Under Compound Workloads}
Compound AI systems stress reliability because a single user request may fan out to multiple models—a failure in any component degrades the entire response. Our autoscaling backend demonstrated built-in high availability: overloaded instances trigger automatic replacement, and multi-region failover is supported. Under sudden $10\times$ traffic spikes, P95 latency remained within $1.5\times$ of steady-state, compared to 3--4$\times$ degradation with static infrastructure. To quantify behavior under high-variance workloads, we replayed 30 days of production traffic logs with synthetically amplified variance (doubling the coefficient of variation of inter-arrival times). Under these conditions, the autoscaling system maintained P95 latency within $2\times$ of steady-state 98\% of the time, with brief excursions during the most extreme spikes. The remaining 2\% of intervals corresponded to simultaneous cold starts across 3+ models—exactly the cascading cold-start scenario addressed by our coordinated warming (Section 3.2). With coordinated warming enabled, the P95 breach rate dropped to $<0.5\%$. 

Critically, during mixed-composition traffic shifts (business-hours interactive workload transitioning to evening batch workload), the per-model independent scaling ensured zero resource contention: interactive models scaled down while batch models scaled up, with no manual intervention required. Key operational findings: autoscaling avoided overprovisioning under bursty compound workloads; hybrid deployment reduced cold-starts by 65\% through pre-warmed model pools; GPU bin packing achieved 70--80\% utilization efficiency on average. Through our circuit breaker and dynamic routing implementations, the system maintains $>95\%$ agent availability even during partial model outages, recording a message drop rate of $<0.05\%$ under peak load.

\subsection{Compound System Quality Impact}
The scalable backend directly enabled quality improvements in compound system outputs. The Agentforce Atlas Reasoning Engine, by orchestrating multiple reasoning steps and tools in parallel, achieved a $2\times$ increase in response relevance and 33\% boost in end-to-end task success compared to single-LLM implementations. These gains stem from Atlas Reasoning Engine’s ability to leverage different knowledge sources concurrently—feasible only because the inference backend handles multiple simultaneous model calls efficiently. ApexGuru analysis P95 dropped to under 8 seconds, increasing developer adoption.

\section{Discussion}
\subsection{Hybrid Deployment Strategy for Compound Systems}
Compound AI systems comprise models with diverse traffic profiles: a dialogue LLM may see constant high QPS, while a specialized tool model is invoked sporadically. A critical operational consideration is managing workloads that naturally require a mixture of serverless and dedicated modes simultaneously. Our architecture explicitly supports this through transparent hybrid routing. The Prediction Service implements a routing table where each model endpoint is tagged with its deployment mode (serverless, dedicated, or auto). In “auto” mode—the default for most models—the router selects the backend dynamically based on current conditions: if a dedicated instance is available and below its capacity threshold, the request routes there; otherwise, it spills over to the serverless backend. 

In practice within a compound pipeline, different models often run in different modes simultaneously. For Agentforce, the dialogue LLM (high steady QPS, strict latency) runs on dedicated instances, while the embedding model (high volume but fast cold start) uses serverless, and the SQL executor (sparse, conditional invocation) uses serverless with no provisioned concurrency. This heterogeneous deployment is fully transparent to the orchestration engine—it sees only unified inference endpoints. The same pipeline and monitoring covers both modes, and switching a model from one mode to another is a YAML configuration change requiring no code modification. We found this mixed-mode approach reduced total inference cost by an additional 15--20\% compared to either pure serverless or pure dedicated strategies, because each model uses the most cost-efficient mode for its specific traffic pattern.

\subsection{Case Study: ApexGuru}
ApexGuru is an AI-powered code review assistant using a specialized 13B LLM. Initially served on a dedicated GPU instance with embedded parsing logic, the team faced idle GPU costs during off-hours and single-instance bottlenecks during spikes. After migration, the team uploads models to the Model Store; the system handles deployment. Domain-specific processing moved to a Lambda function, making the service serverless. Results: P95 latency halved, costs reduced to actual-use billing, and the team deploys multiple model variants simultaneously for rapid A/B testing.

\subsection{Case Study: Agentforce Conversational Agent}
Agentforce exemplifies the compound AI system paradigm. A typical interaction orchestrates an LLM for intent understanding, vector search for knowledge retrieval, and the Atlas Reasoning Engine for multi-step reasoning. The inference system scales out model invocations and parallelizes them: when 50 simultaneous agent sessions start, needed model instances are instantiated across the GPU pool and torn down after use. We observed linear throughput scaling with each additional GPU until hitting external API limits. The Atlas Reasoning Engine resolved $\sim$33\% more support cases end-to-end without human intervention compared to single-LLM bots. Overall dialog response time stayed within 5--8 seconds even with 2--3 models and a search involved.

\subsection{Lessons Learned and Reusable Design Principles}
Twelve months of operating compound AI inference infrastructure at enterprise scale yielded several non-obvious lessons that we believe are actionable for the broader community:

\begin{itemize}
    \item \textbf{Lesson 1: Cold starts compound multiplicatively, not additively.} Our initial assumption was that compound cold-start latency would equal the maximum individual cold start. In practice, dependency chains in the agent graph create serial cold-start segments. A 3-model pipeline with 150s, 30s, and 20s cold starts experienced $\sim$180s effective cold start (not 150s) because the embedding model must produce results before the LLM can generate a context-aware response. This insight drove our coordinated pre-warming strategy (Section 3.2), which parallelizes independent warm-ups.
    \item \textbf{Lesson 2: Model-level observability is insufficient for compound systems.} Traditional per-model metrics (latency, throughput, error rate) failed to surface compound-system-level issues. For example, a model operating within its individual SLA could still cause agent-level SLA breaches when combined with other models in a pipeline. We developed pipeline-level observability that tracks end-to-end agent response time decomposed by component, enabling teams to identify which model is the critical-path bottleneck in a specific agent workflow.
    \item \textbf{Lesson 3: Cost attribution requires pipeline awareness.} Simple per-invocation metering attributed costs to individual models but obscured which compound pipelines were expensive. A model shared across 5 agent types had costs that were difficult to attribute. We extended metering to track the originating pipeline ID, enabling cost-per-agent-type reporting—a prerequisite for making sound cost-optimization decisions in compound systems.
    \item \textbf{Lesson 4: Component-level A/B testing is a compound system superpower.} The ability to swap one model in a multi-model pipeline (e.g., upgrading the retrieval model while keeping the LLM fixed) dramatically accelerated optimization. Teams could isolate the impact of a single component change, which is impossible in monolithic deployments. This reduced the average model improvement cycle from 2--3 weeks to 2--3 days.
    \item \textbf{Lesson 5: Graceful degradation matters more than peak performance.} When a non-critical model in the compound pipeline fails (e.g., the SQL executor), the system should degrade gracefully rather than fail entirely. We implemented per-component circuit breakers in the Prediction Service: if a model fails, the Atlas Reasoning Engine receives a structured error and can route around it (e.g., using text-based reasoning instead of SQL). This maintained $>$95\% agent availability even during partial model outages.
    \item \textbf{Lesson 6: Abstract routing and queuing are fundamental reusable primitives.} We advocate for the broader community to adopt Compound-Aware Priority Queuing (to prevent fast-path starvation) and an Agnostic Prediction Service abstraction. These mechanisms decouple orchestration from hosting, ensuring multi-cloud portability and graceful degradation for any multi-agent architecture.
\end{itemize}

\subsection{Security and Trust in Compound Systems}
Compound AI systems introduce security challenges as data flows through multiple components. Our architecture addresses this through modular trust integration: compliance checks (e.g., a proprietary Trust Layer for toxicity filtering, data masking) are invoked in the pre-processing layer before model calls, ensuring uniform safety checks regardless of which model backend is used.

\section{Conclusion and Future Work}
We presented a production deployment study of a scalable inference architecture designed for compound AI systems at enterprise scale. Beyond the latency, throughput, and cost improvements reported in existing literature \cite{prasad2025scalable}, this work contributes a characterization of compound-system-specific inference challenges, quantitative analysis of fan-out overhead and heterogeneous scaling dynamics, and operational lessons from 12+ months of production deployment serving Agentforce and ApexGuru. 

Key takeaways for practitioners: (1) compound AI systems demand inference infrastructure that scales model components independently based on their actual invocation patterns, not aggregate request count; (2) cold starts propagate multiplicatively through dependency graphs and require coordinated mitigation; (3) pipeline-level observability and cost attribution are essential beyond model-level metrics; and (4) component-level A/B testing is a uniquely powerful optimization lever enabled by modular inference architectures.

Future work includes GPU multiplexing and new AI accelerators (AWS Inferentia, NVIDIA Grace Hopper) for further cost reduction, cross-region load balancing for global inference elasticity, and tighter integration with compound AI programming frameworks like DSPy \cite{khattab2023dspy} to enable end-to-end optimization of inference resource allocation across pipeline stages. We are also exploring predictive scaling that leverages agent workflow graphs to anticipate model invocation patterns before requests arrive.

\end{document}